\documentclass[conference]{IEEEtran}
\usepackage[utf8x]{inputenc} 
\usepackage{graphicx}
\usepackage{mathtools, amsmath,amssymb}

\usepackage[ruled,vlined]{algorithm2e}
\usepackage{caption, subcaption}
\usepackage{todonotes}
\usepackage[para]{footmisc}
\usepackage{booktabs}
\usepackage{multirow}
\usepackage{xcolor,colortbl}
\usepackage{hyperref}
\usepackage{cleveref}
\usepackage{multirow}
\usepackage{rotating}
\usepackage{wrapfig}
\usepackage{lscape}
\usepackage[noadjust]{cite}

\begin{document} % 

\title{What Matters in Explanations: Towards Explainable Fake Review Detection Focusing on Transformers}
\if false
\author{
\IEEEauthorblockN{
Anonymous Authors\\
Authors' Institutes \\
email@correspondence%\\
%\vspace{2em}
}
}
\fi 
\author{
\IEEEauthorblockN{
	Md Shajalal\IEEEauthorrefmark{1}\IEEEauthorrefmark{2},
	Md Atabuzzaman\IEEEauthorrefmark{3}, 
	    Alexander Boden\IEEEauthorrefmark{1}\IEEEauthorrefmark{4},
	Gunnar Stevens\IEEEauthorrefmark{2}\IEEEauthorrefmark{4},
	and Delong Du\IEEEauthorrefmark{2} 
}
    \IEEEauthorblockA{\IEEEauthorrefmark{1} Fraunhofer - Institute for Applied Information Technology FIT, Germany}
    \IEEEauthorblockA{\IEEEauthorrefmark{4} Bonn-Rhein-Sieg University of Applied Sciences, Germany}
    \IEEEauthorblockA{\IEEEauthorrefmark{2} University of Siegen, Germany}
    
    \IEEEauthorblockA{\IEEEauthorrefmark{3} Virginia Tech, VA, USA}
}

\maketitle
\begin{abstract}
    Customers' reviews and feedback play crucial role on electronic commerce~(E-commerce) platforms like Amazon, Zalando, and eBay in influencing other customers' purchasing decisions. However, there is a prevailing concern that sellers often post fake or spam reviews to deceive potential customers and manipulate their opinions about a product. Over the past decade, there has been considerable interest in using machine learning (ML) and deep learning (DL) models to identify such fraudulent reviews. Unfortunately, the decisions made by complex ML and DL models - which often function as \emph{black-boxes} - can be surprising and difficult for general users to comprehend. In this paper, we propose an explainable framework for detecting fake reviews with high precision in identifying fraudulent content with explanations and investigate what information matters most for explaining particular decisions by conducting empirical user evaluation. Initially, we develop fake review detection models using DL and transformer models including XLNet and DistilBERT. We then introduce layer-wise relevance propagation (LRP) technique for generating explanations that can map the contributions of words toward the predicted class. The experimental results on two benchmark fake review detection datasets demonstrate that our predictive models achieve state-of-the-art performance and outperform several existing methods. Furthermore, the empirical user evaluation of the generated explanations concludes which important information needs to be considered in generating explanations in the context of fake review identification.
\end{abstract}


\section{Introduction}
The rapid growth of e-commerce platforms for ordering various products online makes consumers' lives easier, saving potential time and cost for both ends. Issues related to trust and transparency are always of high importance, as they are directly associated with customer satisfaction and the revenue of companies or retailers~\cite{salminen2022creating}. Generally, customers or buyers in e-commerce or service providers tend to check the ratings and reviews of previous customers who have already purchased the products to get an idea of the quality of the targeted products. Users usually prefer to buy products with higher ratings and better reviews from customers. It is evident that companies or retailers sometimes take the opportunity to post fake positive reviews for their products with the objective of making their products appear better. Conversely, opposite scenarios are also visible, where competitors of certain products might post fake negative reviews to portray the products as being of poor quality.

However, identifying fake reviews can benefit both customers and retailers or companies by providing a trusted and transparent e-commerce platform. In the last decade, there has been a significant amount of attention on identifying fake reviews using automated methods with ML and DL-based classifiers. Notable ML methods such as SVM, NB, XGBoost, etc., generally use the TF-IDF or bag-of-words representation of textual reviews~\cite{mir2023online,fontanarava2017feature}. However, these methods are traditional ways of representing text. The semantic representation of text using word embeddings has been employed in almost every natural language processing (NLP) task. Word embeddings can represent the semantic and contextual information of text in a high-dimensional space. With these representations, multiple methods have been proposed using DL-based classifiers, including recurrent neural networks (RNN) and its variants such as LSTM, BiLSTM, GRU, etc~\cite{shajalal2023unveiling,duma2023deep,yu2022graph,paul2021fake,patel2018survey,singh2023deep,deshai2023unmasking,bathla2022intelligent}. After the invention of transformer-based text representations, NLP methods achieved high performance in almost every section. Transformer-based approaches, including BERT and its variants like DistilBERT, mBERT, and RoBERTa, have been used in many text classification tasks~\cite{karim2021deephateexplainer}. Recently, Electra~\cite{clark2020electra}, XLNet~\cite{yang2019xlnet}, GPT, and other large language models (LLMs) have also garnered significant attention in text classification, achieving high performance in numerous NLP tasks.

Generally, DL- and transformer-based approaches have complex architectures and involve a difficult decision-making process in predicting the original class. In the context of fake review detection task, users are typically laypeople with minimal knowledge about predictive models. The decisions might surprise them when they see a particular review detected as fake, but they cannot figure out why it is predicted as a fake review. Recently, explainable artificial intelligence (XAI) has gained significant attention in different fields, including business~\cite{shajalal2022explainable}, bioinformatics~\cite{karim2023explainable}, NLP~\cite{shajalal2023unveiling, karim2021deephateexplainer,arras2017relevant,arras2017explaining}, and more. In the use case mentioned above, XAI comes into play to explain and validate the predictions made by the fake review detector. In this decade, XAI techniques have gained considerable attention in explaining model decisions, allowing AI practitioners and users to understand the reasons behind predictions and improve model performance and decision understanding. Several renowned XAI methods, including SHAP~\cite{lundberg2017unified}, LIME~\cite{ribeiro2016should}, LRP~\cite{arras2017explaining}, Bert-interprete~\cite{ramnath2020towards}, can be applied to explain decisions related to NLP tasks.

The decisions made by complex ML models, which often function as "black boxes," can be surprising and difficult for general users to comprehend. In this research paper, we propose a transparent and efficient framework for detecting fake reviews, enabling high-precision identification of fraudulent content and providing users with explanations to help them understand the predictions. Initially, we develop fake review detection models using cutting-edge transformer models such as XLNet and DistilBERT. We also applied different DL models, including BiLSTM, CNN, CNN-LSTM, and CNN-GRU models for detecting fake reviews. We then introduce LRP~\cite{arras2017relevant,arras2017explaining} technique in fake review detection task to interpret the decisions from DL models and present explanations for individual predictions, highlighting the contributed words for the predicted class. 

We conducted experiments in multiple settings, and experimental results on two benchmark fake review detection datasets demonstrate that our predictive models achieve state-of-the-art performance and outperform several existing methods. Furthermore, our generated explanations can interpret specific decisions, enabling users to understand why a particular review is classified as fake or genuine. The empirical evaluation with 12 human subjects was conducted to examine the effectiveness of the explanations and elicit further requirements in generating explanations in the context of fake review identification. The major contributions in this research are twofold:
\begin{itemize}
    \item We introduced two transformer-based fake review detection models applying DistilBERT and XLNet that demonstrated significantly better performance than DL methods and existing related works. 
    \item Our method is able to explain specific predictions with explanations introducing LRP techniques. The explanations might enable users to make sense of why particular reviews have been predicted as fake.
    \item Our conducted empirical evaluation of the generated explanations with human subjects and the results indicate further requirements in generating explanations for fake content identification tasks.
\end{itemize}

In the rest of the paper, we present the state-of-the-art methods in fake review detection in Section~\ref{LR}. The next section, Section~\ref{OM}, presents our method. The details of the dataset, experimental settings, results on two different datasets, generated explanations with discussion and their empirical evaluation with human-subjects are presented in Section~\ref{ER}. Finally, we conclude our paper with some future directions in Section~\ref{CF}.

\section{Literature Review} \label{LR}
To understand people's strategies for creating and posting fake reviews,~\cite{banerjee2023understanding} conducted an empirical study with participants. Their findings suggested four stages involved in creating fake reviews: information gathering, assimilating information, drafting, and posting the reviews. These stages were identified through qualitative and quantitative methods with fifty-one participants.~\cite{kumar2023exploring} highlighted two challenges in the theoretical grounding and under-researched areas of fake reviews. The first challenge is the lack of a conceptual understanding of the relationship between writing styles and recommendations. The second challenge is the knowledge gap regarding product characteristics. Their empirical investigation, which employed natural language processing (NLP) techniques, revealed latent characteristics of the product corresponding to buying preferences. However, their major findings suggested that the characteristics of fake reviews have no influence on recommending or discouraging the associated product.~\cite{petrescu2023man} analyzed the performance of different automated approaches for detecting deceptive customer reviews, with a focus on evaluating insights provided by multi-modal techniques.

Various classical and deep learning-based text classification models have been utilized, including support vector machines (SVM), k-nearest neighbors (KNN), logistic regression (LR), light gradient boosting model (LGBM)~\cite{rout2018framework,choi2023fake,elmogy2021fake}, long short-term memory network (LSTM), convolutional neural networks (CNN), recurrent neural networks (RNN), and transformers (e.g., BERT and its variants) for identifying the authenticity of online reviews~\cite{duma2023deep,yu2022graph,paul2021fake,patel2018survey,singh2023deep,deshai2023unmasking}.~\cite{deshai2023unmasking} proposed an approach that combines CNN, particle swarm optimization (PSO), and various NLP techniques to identify the credibility and authenticity of online reviews using several datasets. Similarly,~\cite{duma2023deep} presented a deep hybrid model for fake review identification, considering the combination of latent text features, aspect ratings, and overall ratings.~\cite{singh2023deep} applied various classical ML and DL models for identifying fake reviews.~\cite{alsubari2023rule} analyzed online deceptive reviews and proposed a recurrent neural network model for fake review detection, focusing on new features such as authenticity and analytical thinking. They conducted experiments on Amazon and Yelp reviews for electronic products.

Due to the hidden and diverse characteristics of fake reviews, developing a detection framework poses significant challenges.~\cite{shunxiang2023building} proposed a detection framework based on the sentiment intensity of a review and positive unlabelled (PU) learning.~\cite{singhal2023weighted} introduced an ensemble-based learning approach, which balances classes using different sampling techniques (WSEM-S), for modeling the fake review detection task. They applied n-gram models to extract features before applying the model. Their model's performance was compared with conventional ML models, including naive Bayes, XGBoost, KNN, and CNN.~\cite{mir2023online} applied different supervised models, including classical ML models, using BERT representations.~\cite{fontanarava2017feature} also employed supervised models.

A CNN-based fake news detection model was introduced by~\cite{vishwakarma2023framework}, which considers the web-scraped content heading.~\cite{zhang2023deep} proposed a deep learning-based method for fake news detection.~\cite{wang2023vote} applied a voting-based approach to determine whether a review is fake or not, using multiple lists generated by different fake review detection models.~\cite{mohawesh2023explainable} proposed an explainable fake review detection framework using different DL models, including Bi-LSTM, CNN, and DNN. They used Shapley Additive Explanations (SHAP)~\cite{lundberg2017unified} to explain the models. Previously, they addressed the concept drift problem within fake review detection systems~\cite{mohawesh2021analysis,mohawesh2021fake}. With the success of large language models (LLMs) in generating language for various purposes, it is evident that they are also frequently employed to generate artificial product reviews to influence customers' opinions~\cite{salminen2022creating}.~\cite{salminen2022creating} created a fake review dataset using multiple LMs and proposed detection methods for identifying fake reviews.~\cite{bathla2022intelligent} introduced an intelligent fake review detection method using different RNN variants, including CNN and LSTM. They extracted different aspects from the reviews and fed them into the deep learning models. 

Topic modeling techniques were also applied by~\cite{birim2022detecting} to identify fake reviews. They combined review sentiment with other features for fake review identification. A semi-supervised Generative Adversarial Network (AspamGAN) model was proposed by~\cite{jing2022semi}, which incorporates an attention mechanism to address challenges related to the loss of important information due to the relative length of online texts.~\cite{vidanagama2022ontology} introduced an ontology-based sentiment analysis approach that incorporates linguistic features and part-of-speech (POS) for identifying fake reviews. They employed a rule-based classifier based on different extracted features. Different handcrafted features have also been applied for identifying spam online reviews~\cite{shan2021conflicts}.Fang et al.~\cite{fang2020dynamic} proposed a knowledge graph-based method that considers the time and semantic aspects of online customer reviews, as well as multi-source information. Feature-based and content-based classification methods have also been proposed~\cite{barbado2019framework,cardoso2018towards}. Content-based and product-related features have been applied for credibility detection in reviews~\cite{sun2016exploiting}.

From the above-detailed literature review analyzing published papers in the last five years, we can see that none of the methods are explainable except one~\cite{mohawesh2023explainable}. But that one method only applied SHAP value-based explanation for global interpretability. Moreover, the methods proposed to detect fake or deceptive or spam reviews lagged behind the current state-of-the-art methods including transformer-based methods. In this paper, we employ efficient transformers including XLNet and DistilBERT for modeling the fake review detection problem. We compared the performance of different models with baseline deep learning models including LSTM. BiLSTM, CNN, CNN-BiLSTM, and GRU using FastText word-embedding-based text representation. In addition, we applied layer-wise relevance propagation~(LRP) based explanation technique to explain the prediction from our models.

\section{Our Method} \label{OM}
\subsection{DistilBERT Transformer}
DistilBERT~\cite{sanh2019distilbert} is a general purposed distilled version of BERT~\cite{devlin2018bert}. It is 40\% lighter and 60\% faster transformer model than BERT model. However, it retains 97\% language understanding capabilities compared to BERT model. DistilBERT used knowledge distillation technique~\cite{hinton2015distilling} which is a compression mechanism. In this mechanism a compact model, DistilBERT is trained to reproduce the behavior of large model, BERT. To reproduce the behavior of larger models, triple loss was employed in the training phase combining language modeling, distillation and cosine-distance losses. Besides, DistilBERT has the same architecture of the BERT model. As the number of layers have a smaller impact on the computation efficiency, the number of layers is reduced by a factor of 2 in the DistilBERT architecture. The token-type embeddings and pooler are also removed from the design of the compact model. Then DistilBERT is initialized from the BERT by taking one layer out of two and trained on the same corpus as the original BERT model. The technical details of DistilBERT can be found in the paper by~\cite{sanh2019distilbert}.

\subsection{XLNet Transformer}
XLNet~\cite{yang2019xlnet} is a generalized autoregressive approach that utilized the both of autoregressive and autoencoding language modeling. Unlike the usages of the autoregressive models' fixed forward or backward factorization order, XLNet maximizes the log likelihood of a text sequence with respect to all possible permutations of the factorization order to learn the bidirectional context. In contrast to BERT's pretrain-fineture discrepancy, XLNet does not suffer from it. It also use the predicted token's joint probability are factorized with the product rule provided by the autoregressive objectives. XLNet improved its performance by integrating the ideas of Transformer-XL~\cite{dai2019transformer} in its pretraining. As a result, XLNet outperforms BERT on 20 tasks including question answering, sentiment analysis, natural language inference and document ranking, etc. The technical detail of XLNet can be found in the paper by~\cite{yang2019xlnet}.

\subsection{Explaining the prediction}
Inspired by the success of the LRP technique in explaining text classification in different NLP applications, we adopted LRP~\cite{arras2017explaining,arras2017relevant} to explain the prediction to answer ``\textit{why a particular review has been predicted as fake}?'' question. The LRP technique can unveil the black-box deep learning model by back-propagating the output from the output layer up to the input layers through re-distributing the weight in the previous layers. In the end, LRP provides the weight for every input feature and the higher the weight the higher the relevancy of the words towards the predicted class. We then highlight the words based on the provided weight and represent the explanations in terms of highlighted text and word could.

For a deep neural network classification model, LRP re-distributed the weight from the output layers back to the input layers~\cite{shajalal2023unveiling}. Let $i$ be the immediate lower layer and its neurons are denoted by $z_i$. Computationally the relevance messages $R_{i \leftarrow j }$ can be computed as followings~\cite{arras2017explaining}.

\begin{equation}
   R_{i \leftarrow j } = \frac{z_i \cdot w_{ij} + \frac{\epsilon \cdot sign(z_j) + \delta \cdot b_ j} {N} }{z_j + \epsilon \cdot sign(z_j)} \cdot R_j.
\end{equation}

The total number of neurons in the layer $i$ is denoted as $N$ and $\epsilon$ is the stabilizer, a small positive real number (i.e., 0.001). By summing up all the relevance scores of the neuron in $z_i$ in layer $i$, we can obtain the relevance in layer $i$, $R_i = \sum_i R_{i\leftarrow j}$. $\delta $ can be either 0 or 1 (we use $\delta =1$)~\cite{arras2017explaining,karim2021deephateexplainer,shajalal2023unveiling}. 

\section{Experiments}\label{ER}
\subsection{Dataset}
\noindent\textbf{Fake Review Dataset:} This fake review dataset contains 40000 reviews in total. Among them, 50\% reviews were originally written by humans~(i.e., reviews collected from Amazon). The rest of the reviews are fake, generated by two different language models including ULMFit~(Universal Language model Fine-tuning) and GPT-2~\cite{salminen2022creating}. The reviews are for products from 10 different categories~\cite{salminen2022creating}. The categories are \textit{Books, clothing, shoes and Jewelry, Electronics, Home and Kitchen, Kindle, Movies and TV, Pet Supplies, Sports and Outdoors, Tools and Home Improvements, and Toys and Games}. %The distribution of the number of reviews among different categories is depicted in Fig.~\ref{fakeReviewDistribution}.

%\begin{figure*}[!htb]
%    \centering
%    \includegraphics[width = 0.95\linewidth]{figures/data_distribution.pdf}
%    \caption{Distribution of samples among different categories.}
%    \label{fakeReviewDistribution}
%\end{figure*}

\noindent\textbf{Yelp Review Dataset:} We conducted experiments with another fake review dataset named \textit{Yelp Review Dataset}. Compared to the previous one, this dataset is quite big and consists of more than 682K reviews and the distribution is quite imbalanced. The dataset is accessible at Kaggle\footnote{https://www.kaggle.com/datasets/abidmeeraj/yelp-labelled-dataset/data}. %To illustrate the imbalanced between two different classes, we present the samples distribution in Fig.~\ref{YelpFakeReviewDistribution}.
%\begin{figure}[!htb]
%    \centering
%    \includegraphics[width = 0.95\linewidth]{figures/yelp_data_distribution.pdf}
%    \caption{Distribution of samples among classes.}
%    \label{YelpFakeReviewDistribution}
%\end{figure}
%\vspace{-1em}
\subsection{Experimental Settings}
To assess the efficiency of our introduced fake review detection methods, we conducted a comprehensive set of experiments using two distinct fake review datasets. We initially applied an ensemble machine learning (ML) method, employing a majority voting technique on four different ML models: Support Vector Machine (SVM), Decision Tree (DT), Random Forest (RF), and XGBoost. Subsequently, we explored four distinct deep learning models, namely BiLSTM, CNN, CNN-LSTM, and CNN-GRU. In the BiLSTM model, we incorporated an embedding layer followed by a Spatial dropout layer. This was followed by two bidirectional LSTM layers, each consisting of 64 and 32 BiLSTM units, along with a dropout layer after each. The model then incorporated a fully connected layer and concluded with an output layer featuring the \textit{sigmoid} activation function.

Similarly, the CNN model began with an embedding layer, followed by a Spatial dropout layer. A convolutional layer was introduced, followed by a dropout layer, and finally, two fully connected layers. The CNN-LSTM model represented a hybrid combination of CNN and LSTM layers. It started with an embedding layer, followed by a CNN layer, an LSTM layer, and two fully connected layers. The architecture for the CNN-GRU model closely resembled that of the CNN-LSTM model, with the key distinction being the replacement of LSTM layers with GRU layers. For our transformer-based approaches, we employed DistilBERT and XLNet transformers for the purpose of detecting fake reviews. Since there is a high chance of vocabulary mismatch problem, we employed \textit{FastText}, a character embedding-based pre-trained word embedding model, to represent the reviews semantically for deep learning models. For transformer models, we employed \textit{distilbert-base-uncased} and \textit{xlnet-base-cased} for DistilBERT- and XLNet-based classification models, respectively. For all experiments, we split both datasets into train, test, and validation sets in 70\%, 15\%, and 15\%, respectively.

\begin{table*}[!htb]
    \centering
    \caption{The performance of different methods compared to baselines on Fake Review Dataset.}
    %\vspace{-0.5em}
    \begin{tabular}{|c|c|c|c|c|}
        \hline
        \textbf{Type} & \textbf{Model} & \textbf{Accuracy} & \textbf{Precision} & \textbf{F1Score} \\ \hline
        \multirow{1}{*}{Baseline}
         & EnsembleML & 0.8425& 0.9147& 0.9014 \\ \hline

         \multirow{4}{*}{Deep Learning}& BiLSTM& 0.9556 & 0.9750   & 0.9466  \\ \cline{2-5}
         & CNN& 0.9252 & 0.9268 & 0.9259 \\ \cline{2-5}
         & CNN-LSTM& 0.9486  & 0.9454  & 0.9493 \\ \cline{2-5}
         & CNN-GRU& 0.9476 & 0.9751 & 0.9466 \\ \hline

         \multirow{2}{*}{Transformers}
         %& BERT& & &  \\ \cline{2-5}
         & DistilBert& \textbf{0.9592} & \textbf{0.9906}& \textbf{0.9821} \\ \cline{2-5}
         & XLNet& 0.9580  & 0.9887 &  0.9779\\ \hline
          
    \end{tabular}
    \vspace{-1em}
    \label{FakeReviewPerformance}
\end{table*}

\begin{table}[!htb]
    \centering
    \caption{The performance of the fake review detection method for different product categories.}
    %\vspace{-0.5em}
    \begin{tabular}{|c|c|c|c|}    
    \hline    \textbf{Category}&\textbf{Accuracy}&\textbf{Precision}&\textbf{F1Score} \\ \hline
        Home & 0.9557  & 0.9786  & 0.9689 \\ \hline
        Sports & 0.9544 & 0.9750  & 0.9686 \\ \hline
        Electronics & 0.9461  & 0.9807  & 0.9479 \\ \hline
        Movies & 0.9499  & 0.9689  & 0.9631 \\ \hline
        Tools & 0.9508  & 0.9707  & \textbf{0.9711} \\ \hline
        Pet & \textbf{0.9612}  & \textbf{0.9833}  & 0.9705 \\ \hline
        Kindle & 0.9461  & 0.9795  & 0.9508 \\ \hline
        Books & 0.9439  & 0.9772  & 0.9459 \\ \hline
        Toys & 0.9341  & 0.9708  & 0.9325 \\ \hline
        Clothing & 0.9351  & 0.9802  & 0.9295  \\ \hline
    \end{tabular}
    \vspace{-1em}
    \label{Performance_Category}
\end{table}

\subsection{Experimental Results}
\subsubsection{Performance on Fake Review Dataset} The performance of different fake review detection models on Fake Review dataset~\cite{salminen2022creating} is presented in Table~\ref{FakeReviewPerformance} in terms of multiple evaluation metrics. Among four different deep learning models, BiLSTM performs better in terms of accuracy~(0.9556) and F1-Score~(0.9466). We can also see that CNN-GRU performs equally compared to BiLSTM in terms of F1-Score and Precision which is almost the same. However, the other two DL models CNN and CNN-LSTM also achieved consistent and effective performance. In terms of all evaluation metrics, our proposed two transformer-based fake review detection models achieved significantly higher accuracy~(0.9592), precision~(0.9906), and F1-Score~(0.9821) among all employed models. The performance difference between XLNet and DistilBERT is not significant and it is only a 1\% difference in terms of precision. DistilBERT achieved more than 4\% performance gain in terms of F1-Score. 

To illustrate the performance of our best method~(DistilBERT) in identifying fake reviews, we present the performance across different categories. Table~\ref{Performance_Category} presents the performance for reviews in different product categories. Identifying fake reviews for \textit{pet supplies} category, DistilBERT achieved 96\% accuracy and 98\% precision compared to the performance on other categories. On the other hand, it achieved higher F1-Score of 0.9711, which is higher across all categories. 
\begin{table}[!h]
    \centering
    \caption{Performance comparison with existing method on fake Review Dataset. The model \textit{OpenAI, NBSVM and fakeRoBERTa} are proposed by~\cite{salminen2022creating}.}
    %\vspace{-0.5em}
    \begin{tabular}{|c|c|c|c|}  
    \hline
    \textbf{Method} & \textbf{Precision} &\textbf{Recall} &\textbf{F1Score} \\ \hline
       OpenAI & 0.83 & 0.82 & 0.82 \\ \hline
       NBSVM & 0.95&   0.95 & 0.95\\ \hline
       fakeRoBERTa  & 0.97  & 0.97  & 0.97  \\ \hline
       \textbf{DistilBert} & \textbf{0.99}& \textbf{0.97} & \textbf{0.98} \\ \hline
       \textbf{XLNet} & 0.98 & 0.97 & 0.97\\ \hline
    \end{tabular}
    \vspace{-3em}
    \label{comparisonSOTA}
\end{table}

We compared the performance of our transformer-based models with state-of-the-art methods on the same dataset by ~\cite{salminen2022creating}. They applied three different classification models to identify fake reviews. First, they trained support vector machine-enabled classifier with Naive Bayes Feature~(NBSVM). Then, an OpenAI model is specifically developed and applied for fake review detection. The OpenAI model is based on the idea of Roboustly Optimized BERT Pretraining Approach~(RoBERTa) with fine-tuning. Finally, inspired by the performance on OpenAI, they designed a RoBERTa-based customized model called \textit{fakeRoBERTa} as their final model. The comparison presented in Table~\ref{comparisonSOTA} shows that our both transformer models XLNet and DistilBERT achieved significantly better performance in terms of precision and F1-score. In terms of recall, their performance is quite consistent. However, the performance compared to the results of a wide range of experiment settings and existing methods, our introduced DistilBERT model demonstrated a new state-of-the-art performance in fake review detection tasks. 

\begin{table*}[!htb]
    \centering
    \caption{The performance of different methods compared to baselines on Yelp Review Dataset.}
    %\vspace{-0.5em}
    \begin{tabular}{|p{1.7cm}|c|c|c|c|c|}
        \hline
        \textbf{Type} & \textbf{Model} & \textbf{Accuracy} & \textbf{Precision} & \textbf{F1Score} & \textbf{AUC}\\ \hline
        \multirow{1}{*}{Baseline}
         & EnsembleML & 0.7848& 0.7795& 0.8156& 0.6271   \\ \hline
         
         \multirow{4}{*}{Deep Learning }
         & BiLSTM& 0.8947 & 0.8985  & 0.9444 & 0.7236 \\ \cline{2-6}
         & CNN& 0.8961 & 0.8966  & 0.9451& 0.7330  \\ \cline{2-6}
         & CNN-LSTM& 0.8842 & 0.9007 & 0.9380&0.6780  \\ \cline{2-6}
         & CNN-GRU& 0.8964 & 0.8978  & 0.9452& 0.7249 \\ \hline
         
        \multirow{3}{*}{Transformers}
         %& BERT & & & & &  \\ \cline{2-6}
         & DistilBert& 0.9235& \textbf{0.9326} & 0.9595 & 0.7958 \\ \cline{2-6}
         & XLNet& \textbf{0.9349} & 0.9278& \textbf{0.9654} & \textbf{0.8044} \\ \hline
          
    \end{tabular}
    \vspace{-1em}
    \label{YelpPerformance}
\end{table*}

\subsubsection{Performance on Yelp Review Dataset:} 
We also carried out experiments on another dataset to demonstrate performance consistency. As we noted earlier the Yelp dataset is quite imbalanced and the number of majority samples is way larger than the minority samples, we measure the performance in terms of area under curve measure along with accuracy, precision, and F1-Score. We present the performance for the Yelp dataset in table~\ref{YelpPerformance}. The table summarized that transformers-based classification models here also performed better than the deep learning models and ensemble ML model. Unlike the performance in the previous dataset, XLNet achieved higher accuracy, F1-Score and AUC compared to the DistilBERT-based classifier. But for the other measure, in terms of precision , DistilBERT performed better. However, the performance difference is not that big but compared to the deep learning-based methods, both DistilBERT and XLNet outperformed significantly with a way higher AUC. AUC is considered one of the best evaluation metrics to measure the performance when the dataset is imbalanced. 

Overall, the performance on this dataset is lower than on the previous dataset. There are several probable reasons. One is the size of the dataset, the Yelp dataset is significantly larger than the fake review dataset and reviews are written by human. However, in the Fake review dataset, the fake reviews are generated by the large language models~(LLM). Additionally, the Yelp data is considerably imbalanced. Since the reviews are generated by LLM, the transformer-based classification models might recognize the review patterns better than the reviews written by humans. However, considering the performance of a wide range of experiments on two different datasets, we can conclude that DistilBERT and XLNet achieved new state-of-the-art results in identifying fake reviews, both for human and machine-generated fake reviews.

\subsection{Explainability of the predictions}
%In this section, we demonstrated the explanations for particular predictions based on the provided weight for each word by LRP re-distribution techniques. 
We present explanations provided by LRP technique using highlighted text and word cloud where the color intensity in highlighted text and size of the words  \textit{WordCloud} represent the degree of relevancy towards the class. We demonstrated explanations for two predictions from each dataset. The first considered review is for a book. It is a fake review generated by a transformer-based language model and our model also predicted it correctly, as fake. The review is as follows:\\
%\vspace{-1em}

\noindent\textbf{Review 1:}\textit{ ``First, let me say I'm an avid reader and this is a book that I read as a child. I had to read it before. I could have a chance to take it to college. I still enjoy reading it as a kid. This book is still one of my favorite books. I have read the book over and over again and it is a must read. I just can't put it down. The only reason I gave it five stars is because I want to read more about the characters. I liked the way they interacted with the kids. I loved their reactions to.''}

\begin{figure*}[!htb]
    \centering
    \includegraphics[width = 0.65\linewidth]{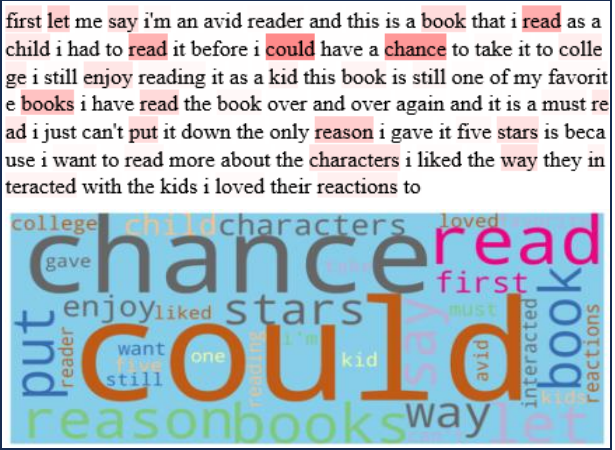}
    \caption{Explanation with highlighting relevant words for a predicted fake review.}
    \vspace{-1.2em}
    \label{FakeReview1}
\end{figure*}

The explanation is depicted in Fig.~\ref{FakeReview1}. We can see the highlighted words are related to the predicted class. The highlighted text and word cloud also show that words such as \textit{read, chance, enjoy, liked, loved stars} are some most relevant for the prediction. We have a closer look at the review text, it is a review with exaggeratedly praises the book. The highlighted words are used for exaggerated praise. Let's consider another fake review for the clothing category:\\
%\vspace{-.5em}

\textbf{Review 2:}\textit{ ``I actually have a review here on the site about these gloves. These gloves are awesome and i highly recommend them. I am a 32-year-old man who works at a large company and have had no issues with these gloves. I have had no issues with the gloves falling apart or breaking. They are the most comfortable I have ever worn, and I am so happy with them. I recommend them to anyone looking for a glove that will last a long time. I am very happy with my purchase and would recommend these gloves to anyone looking for a great glove. These are super lightweight and lightweight. I have used them for a couple of months, and they are still working great. I would definitely recommend them to anyone looking for a good quality product. I've had no problems with them slipping or sliding my only complaint is that I haven't used them yet, but they are very well made and have been used in my gym.''
}

\begin{figure*}[!htb]
    \centering
    \includegraphics[width = 0.65\linewidth]{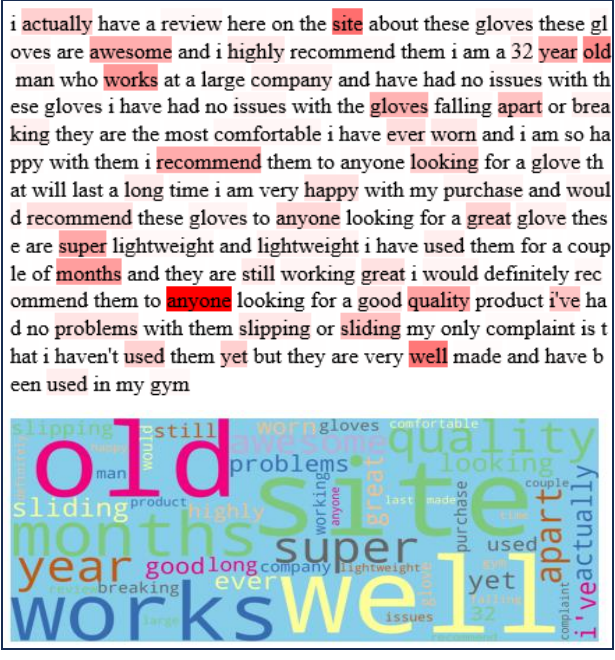}
    \caption{Explanation with highlighting relevant words for a predicted fake review.}
    %\vspace{-1.2em}
    \label{FakeReview2}
\end{figure*}

The explanation for this review is illustrated in Fig.~\ref{FakeReview2}. This is a long review for \textit{gloves} and if we have a closer look on the the text, we can see that it also has redundant praise~(i.e., recommends the product several times and claims that the gloves were the best the user experienced ever). However, the LRP-enabled explanation identified words that indicated why it is classified as fake. The relevant words are \textit{awesome, recommend, anyone, quality}, etc.

\begin{figure*}[!htb]
    \centering
    \includegraphics[width = 0.65\linewidth]{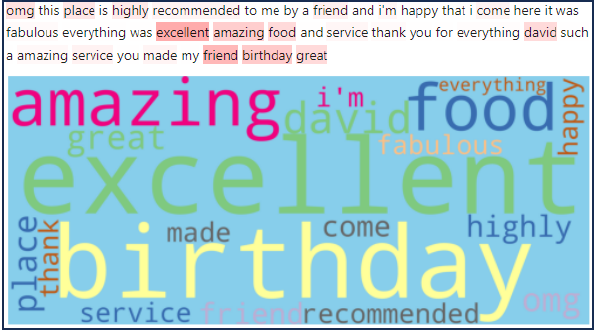}
    
    \caption{Explanation with highlighting relevant words for a predicted fake review for Yelp Dataset.}
    \vspace{-1.6em}
    \label{yelpfakereview1}
\end{figure*}

In the Yelp dataset, the review text is not fine-grained since all the reviews, both real and fake, are written by humans. The grammatical quality is not similar to the previous dataset. However, let's look at the explanation of a review in Fig.~\ref{yelpfakereview1}. This is a review for a restaurant. \\
%\vspace{-.5em}

\textbf{Review 3:}\textit{ ``Omg this place is highly recommended to me by a friend and I'm happy that I come here. It was fabulous. Everything was excellent. Amazing food and service thank you for everything David. Such a amazing service. You made my friend birthday great.''}

\begin{figure*}[!htb]
    \centering
    \includegraphics[width = 0.65\linewidth]{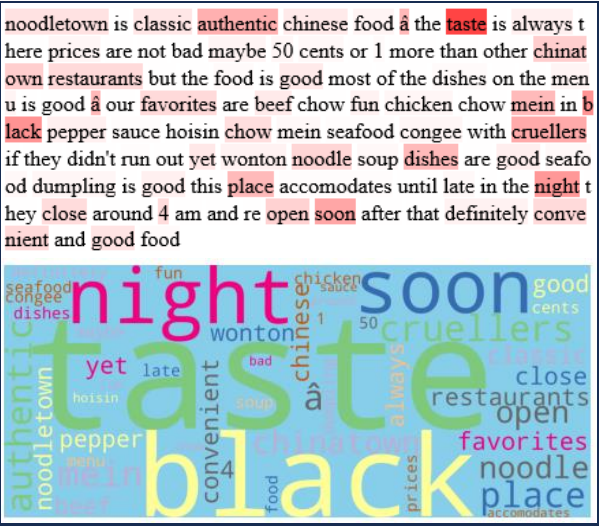}
    \caption{Explanation with highlighting relevant words for a predicted fake review for Yelp Dataset.}
    \vspace{-1.7em}
    \label{yelfakereview2}
\end{figure*}

We can see from the figure that the highlighted words including \textit{amazing, fabulous, great, service, highly, recommended} etc. play bigger roles in helping the classifier to decide it is a fake review. For another review which is quite longer than the previous one. Fig.~\ref{yelfakereview2} presents that the responsible words for which the deep learning models decide it as a fake review are \textit{authentic, taste, favorites, dishes, open,  place, } etc.\\
%\vspace{-.5em}

\noindent\textbf{Review 4:}\textit{``NoodleTown is classic authentic chinese food. â The taste is always there prices are not bad maybe 50 cents or 1 more than other Chinatown restaurants but the food is good. Most of the dishes on the menu is good. â Our favorites are beef Chowfun, Chicken Chowmein in black pepper, sauce hoisin chowmein, seafood Congee with Cruellers if they didn't run out yet. Wonton Noodle, Soup dishes are good seafood. Dumpling is good this place accomodates until late in the night. They close around 4 am and re open soon after that definitely convenient and good food.''}

\normalsize 

\subsection{Empirical User Evaluation}
To evaluate the effectiveness of the LRP-generated explanations highlighting the important relevant words to the predicted class, we conducted an empirical user study with 12 human subjects. The subjects are studying master's in business informatics. We first give then an overview of how our transformer-based model predicts the authenticity of the review. Then we provide them with a simple demo about the explanations and what those highlighted words mean. 

We provided them three reviews (Review 1, Review 2, and Review 3) and asked them to score how authentic the reviews were. All three reviews were fake but we have not told them. Because we wanted to observe how they identify and what are the logic behind. We also provided the details about the products for which the reviews were posted. They were asked to put score for each review, and the score ranges from one star (*) to five star (*****). The highest value 5 (*****) indicates that the corresponding review is original, while the lowest value 1 (*) indicates the review is fake. The participants first score each review after carefully reading the reviews without the generated explanations. 

We then provided them with the LRP-generated explanations (Fig., \ref{FakeReview1},\ref{FakeReview2},\ref{yelpfakereview1}) for each reviews, respectively. We then instructed the participant to look at the explanations and re-score the reviews whether their assumptions changed after perceiving the explanations. We denote two scores before and after the explanations as \textit{score 1} and \textit{score 2}, respectively.

We then discuss with every participant why they think a particular review is original or fake. What are the reasons and rationale behind their scores? We also asked them about the efficiency of the generated explanations, and how they help the participants to decide on the authenticity of the reviews.
\begin{table}[!htb]
    \centering
    \caption{The user evaluations whether the reviews are fake or real, with and without explanations.}
    \begin{tabular}{|c|l|l|l|l|l|l|}
        \hline
         \multirow{2}{*}{\textbf{Subject}} & \multicolumn{2}{c}{\textbf{Review 1}} & \multicolumn{2}{c}{\textbf{Review 2}} & \multicolumn{2}{c|}{\textbf{Review 3}} \\ \cline{2-7}
         &Score1&Score2&Score1&Score2&Score1&Score2 \\ \hline
         1 & **&** & ****&***   & **&**\\ \hline
         2 & **&** & ****&**** &  ****&****\\ \hline
         3 & *&* & *&* &**&**\\ \hline
         4 & ***&** & *&* &****&***\\ \hline
         5 & *&* & *&* &****&****\\ \hline
         6 & **&** & **&*** &*****&*****\\ \hline
         7 & **&** & *&** &****&****\\ \hline
         8 & *&* & *&** &*****&**\\ \hline
         9 & *&* & *&** &***&****\\ \hline
        10 & *&* & ***&*** &**&**\\ \hline
        11 & *&* & *&** &***&***\\ \hline
        12 & ***&* & **&*** &*****&*****\\ \hline
         
    \end{tabular}
    
    \label{empirical_evaluation}
    %\vspace{-1em}
\end{table}

Table~\ref{empirical_evaluation} represents the evaluation of the participants on whether those three reviews are original or fake. We can see that all participants thought that review 1 was fake except subjects 4 and 12. They provide three stars out of five, which concludes it is somewhat original. However, they changed their decision after having the explanations bu putting two and one star, respectively. For review 2, except for participants 1 and 2, everyone considered the review to be fake. Interestingly, review 3 were considered solely as original by the majority of the participants. However, after considering the explanations generated by the LRP-enabled explainability technique, two participants (subject 4 and 8) changed their decision by decreasing the mark.

\noindent\textbf{Discussion on participants' opinion:} We had a detailed discussion with each participant on how they came up with the decision whether a particular review was fake or original. For example, we asked the subject about the review 3. He said the following: 

\textit{Subject 2: ``The third review, because it was for me it was the most realistic. There was the name inside. So he seems to know the guy who's doing it and it's pretty, and it's really short.''}

He thought it was short and he believed in the text because it has a name. However, we asked him, what matters in predicting the review whether it is real or fake. He replied, its more about \textit{linguistic form} (i.e., meaning grammatical structure and tone), not individual word.

\textit{subject 2: ``Yeah, and the second it's, uh, more about the words. And in the first, first, it's more about the linguistic form to me.''}

Similarly, Subject 3 also thought that highlighting relevant words as an explanation might not make sense in explaining review identification whether it is fake or real. It's about the whole text. He added, for the second review, based on the repetitive texts he identified review 2 as fake.

\textit{Subject 3: Uh, the second one is, I also think it's completely made up by AI, um, because it's very repetitive and, uh, uh, some sentences you just read and you think no human would write like this. Um, and then the third one to me also was the most realistic one because it is kind of short. It's very, it kind of seems authentic in terms of like the excitement. }

\textit{Subject 3: ``No, to me, it's not the singular words. To me, it really is the structure and the whole like, the thing as a whole that, um, makes it seem like it's AI generated''}

Subjects 4 and 5 provided their insight about whether our generated explanations make sense to understand the decision. They both thought that the current form of explanation might help to some degree to comprehend the decision, 2 out of 10.

\textit{subject 4: : ``I think it might, it might make sense to some degree. But as, uh, my colleague just said, it's more about the, the overall. Two out of 10''}

\textit{Subject 5: ``No, no. Not from my part. I have to reread that, like out of 10. Uh, like two or three.''}

Subject 6 has found something very interesting in review 2, for example, information like age, and jobs are not relevant and these are not commonly used in review. He also identified that this is a very long review, generally, people are too lazy to write.

\textit{Subject 6: ``Because no matter, um, his age or his, um, job and something like this or for buying gloves, um, and also it's, um, too long. And I guess, um, people are, most people are lazy to write this kind of message. Yeah.''}

Grammatical information is identified as important to understand whether the review is fake or real by subject 7. He considered the more the number of adjectives that exist in the review, the more realistic the review is.

\textit{Subject 7: ``No, just any adjective. So for example, the ones that I have rated the, the, the realest, have more objectives than, than the other ones. So it could be just, um, your personal opinion, it's not about.''}

He also thought the individual word might have some importance towards certain classes, but it should be the whole context of the review. 

\textit{Subject 7: ``For me, for me, they didn't really help me to find out if they are or not real. Uh, I think it's more like a context thing. Only, I mean for me the word has, has to, um, it's okay. It was the, the same.''}

Interestingly Subject 8 found our generated explanations are effective. Before accessing the explanations, subject 8 provided review 3 as 5 starts, meaning a fully real review. But after he went through the generated explanations, he thought this was also a fake review. Though it has several good adjectives, but he thinks these are the reasons to be fake, contradictory to subject 7. 

\textit{Subject 8: ``So, the third one actually was from me, at the first, um, I gave them five stars. So that it's likely to happen because it's short. I think in real life everyone just give short recommendations and not long recommendations. But then after reading the AI, um, explanation, yes. Um, reading the words excellent, amazing, great. I also think that it's, it's not a real review.''}

In summary, the explanations generated by LRP technique to highlighting important relevant words in context of fake review identification can make general users sense in minimum scale. On contrary, for some application areas, for example, sentiment classification~\cite{arras2017relevant} and hate speech recognition~\cite{karim2021deephateexplainer}, where LRP-based explanations are quite good to understand the reason behind the prediction. One of the main reasons identified through the empirical user study why explanations highlighting relevant words is the use of similar words in both fake and original review. For both positive and negative reviews, we observed similar adjectives or other praising or criticizing words are used in both fake and original reviews. For example, in sentiment analysis task, there are some terms or negation elements that are used for specific positive and negative class~\cite{arras2017relevant}. For another example,  patent classification task~\cite{shajalal2023unveiling}, terms related to specific scientific fields are used in the patent text. Rather, in the context of fake review identification task, grammatical structure of the sentence, tone and overall context matters most in explaining the decisions.
\section{Conclusion and Future Direction}\label{CF}
In this paper, we proposed transparent and interpretable fake review detection framework applying transformer models including DistilBERT and XLNet. We also apply multiple deep learning models including BiLSTM, CNN, CNN-LSTM, and CNN-BiLSTM for modeling the fake review detection task. Then, we adopted LRP technique to open the the black-box deep learning model. The LRP can explain why a particular prediction has been made. We conducted experiments in multiple settings and applied our models to two different benchmark datasets. Based on the experimental results, we demonstrated that our proposed DistilBERT- and XLNet-based fake review detection models significantly outperformed other ensemble ML and DL models. Compared to the previously known related methods, our method also outperformed NBSVM, OpenAI, and fakeRoBERTa methods on the same dataset. In the end, we demonstrated explanations provided by our adopted LRP technique for multiple example reviews for different categories. The empirical user evaluation with human subjects indicates further requirements to generate and present the explanations for any specific decision.
%Our method can identify relevant responsible words that make sense as to why the predictive model decides a review is fake. 

In the future, we are planning to have an empirical study to measure the quality of the generated explanations. Further, it would be interesting to consider the elicited requirements and findings from user evaluation for explanation generation.
\bibliographystyle{IEEEtran}
\bibliography{references.bib}
\newpage

\if false
\section*{Appendix}

Person 1: Use some times to tell us what you, what are the main reasons, like just tell, tell us like each paper record.\\

Person 2: Yeah. Yeah, about the, the bombs, louder please. I think it's a AI about the AI checker. Because I think the words are not the same in the normal users can use these words. For example, I use this words when I make a review. Uh, about the second review, about the second review, uh, I think the second review is a real review, not made by AI. And I'm not agree with him because the, uh, it's show us that he's a 32 year old man and they told everything about his life. And it's, I think it's realistic. And the reviews three, uh, it's not real, I think because the, but I don't know about the David, only the world David. No. I can't understand which one is it, but I agree with the, uh, AI checker here for the review three. Okay. Thanks.\\

Person 1: You can put it there and you can go then.\\

Person 2: Go catch your bus.\\

Person 1: Anyone else needs to leave early? You guys need to leave earlier? No, you're not allowed to. Um, but yeah, go ahead. Tell us like, just, just like the way he did.\\

Person 3: Ladies first, go.\\

Person 1: Yeah, go.\\

Person 3: Say it louder, I think. Uh, I would disagree with the third, because for me the, the third review? The third review, because it was for me it was the most realistic. There was the name inside. So he seems to know the guy who's doing it and it's pretty, and it's really short. I hate to, to write really long reviews and I think most of the people are the same.\\

Person 1: So according to your experience, it should be real. Yeah, I think. Okay.\\

Person 3: Um, and I think the first one is, uh, I generated because the sentence sometimes make no sense and they were partly contradictory. Um, and about the second one, I wasn't really sure. It was a, could be or not because there was a lot of words, a lot of really strong words. Um, but as you say, 42 year old, or 32 year old, maybe, yeah, it's a lot of, yeah.\\

Person 1: What do, what do you think, uh, these reviews is, um, like, you are talking, like for, yeah, like, uh, you were talking some reviews are maybe not real. What is your, uh, like thought, uh, is it, uh, because of the linguistics forms or the words or something else? Like, maybe, yeah, it's.\\

Person 3: Yeah, and the second it's, uh, more about the words. And in the first, first, it's more about the linguistic form to me.\\

Person 1: Second means the last, uh, second review you're talking about, okay? Yeah, second review. Okay. Hm, okay. Yeah, tell me about yourself.\\

Person 4: Okay. Uh, review three sounds most realistic to me as well. Uh, because it is as she already said, um, short sentences. And they described for me most accurately an experience where the person actually thinks, okay, this was my experience. I thank the person called, what was his name? Dan or whatever. Um, David, uh, so yeah, sounds most realistic to me. Um, review two were some, uh, and also review three had spelling mistakes. And I don't think that AI would make spelling mistakes since it's trained on like, words without spelling mistakes. Uh, review two had some really weird passages where for example, lightweight and lightweight, these are super lightweight and lightweight. Um, which seem very suspicious. Um, and then review two and one have both in common that there are paragraphs with basically no input. They just say, "Oh, it's very nice and could recommend it because it's very nice." And that's not giving an actual review about something because he's not commenting on what is good about the product or bad about the product. He's just saying, "It's nice. I can recommend that it's nice."\\

Person 1: Okay. Yeah. Yeah, uh, what do you think for all three reviews? Well, um, review one I think has, uh, as my colleague said a lot of, uh, meaningless paragraphs, is very confusing sometimes, uh, contradicting as well. And just the whole review doesn't make any sense to me. So, I, that one to me was entirely, um, AI made up. Uh, the second one is, I also think it's completely made up by AI, um, because it's very repetitive and, uh, uh, some sentences you just read and you think no human would write like this. Um, and then the third one to me also was the most realistic one because it is kind of short. It's very, it kind of seems authentic in terms of like the excitement. Um, whereas the other ones, when there was some excitement, it felt like, oh, this is a great product. I love it. And then, oh, this is so great. It didn't feel as authentic as the third one. Um, so yeah.\\

Person 1: Yeah. So, what is your thought on the explanations that, uh, for example, for this one, like this is review two. You said that you also think that this is a fake review, and, um, uh, the ML model highlighted some word, um, and do you, do you agree with such words that, uh, from, like it's because of such words is used here. That's why it's, it's fake or?\\

Person 4: No, to me, it's not the singular words. To me, it really is the structure and the whole like, the thing as a whole that, um, makes it seem like it's AI generated. And like, sometimes the repetition of words, um, but not, I wouldn't single out any singular word and be like, oh, this, this is an indication to me that it is AI generated. Um, if I look at here, the second one, uh, the word is anyone. And if I say, oh, this product could be used by anyone, then I think that's a pretty normal sentence that could be used in an authentic review as well. So, I, I don't disagree, uh, I don't agree with the AI.\\

Person 1: Thank you. Thank you. Who's next?\\

Person 5: I agree to my colleagues that the last one seems to be the only realistic one, because the other ones are overly long and have some kind of unnecessary information in them, because there is no reason to write so many repetitive sentences about the same, uh, topic. And, uh, I agree as well that I don't think that you can, um, do you think the, uh, the explanation by AI, uh, make some sense to you or help something you to make it?\\

Person 1: I think it might, it might make sense to some degree. But as, uh, my colleague just said, it's more about the, the overall.\\

Person 5: If I may ask, uh, if I would like to, uh, request you to give some rate that how, how much it helps, like out of 10, maybe two, I don't know. Okay, okay. I don't think it's too much. It looks like random words. It's hard to say. I don't know. Yeah. Yeah, I understand. Thank you so much.\\

Person 1: And, uh, what about you? Um, I think.\\

Person 6: Possible, please say it louder.\\

Person 1: Okay.\\

Person 6: So, um.\\

Person 1: Yeah, good.\\

Person 6: The first review, I think. I don't know. I, I'm really not sure about every, anyone of them because I'm a little bit.\\

Person 1: Yeah, but maybe, you, you rate it, um, okay.\\

Person 6: Yeah.\\

Person 1: As fake, okay?\\

Person 6: Yeah, I rate it as fake, but I'm not sure because Amazon sends, send us mails to give review to some products and some people click on it and write some random shit about it. So I don't know if it's real or fake, so I, so, but for me none of these reviews are, uh, I wouldn't decide on a product for none of these reviews because there is no point in it. And I don't, uh, give extra information on a product which I want to know about the product to buy or not buy the product. Maybe there is some point of it where I don't want to use it and yeah, you cannot, you cannot see anything. So, it's not conclusive from your part, right? Okay. So, you, you cannot decide that whether these are the real or fake or, uh, uh, do you think the explanations or the highlighted work and the world cloud can help some in, in some, uh, like?\\

Person 6: No, no. Not from my part.\\

Person 1: Yeah, maybe a bit, but, uh, yeah.\\

Person 6: I have to reread that, like out of 10. Uh, like two or three.\\

Person 1: Two or three, okay? Yeah. Thank you so much. Yeah, what about you?\\

Person 7: Uh, I also totally agree with other, uh, colleagues said, uh, about first one, um, uh, I am pretty sure that I was pretty sure that's fake, because it's a kind of exaggerated message and it's repeated some specific points too many. Um, such a, um, for example, "I must read this book" or, um, "I just can't put it down" and this kind of things. Uh, about second, um, comment, um, I again, uh, totally sure that it's fake, because, um, it has lots of unrelated information. Because no matter, um, his age or his, um, job and something like this or for buying gloves, um, and also it's, um, too long. And I guess, um, people are, most people are lazy to write this kind of message. Yeah.\\

Person 1: Maybe somebody is working overwhelming with the product and write a long reviews. That can happen sometimes. Yeah.\\

Person 7: Yeah. It's not usual, so, uh, but he's right. Maybe, yeah, yeah, yeah. I agree.\\

Person 1: Something that I found out or not found out, but I thought, uh, maybe, those paragraphs that have more adjectives, maybe are more real, maybe? Or is it that like, something.\\

Person 7: So the more, uh, there's a number of objectives you have in the paper, uh, in the paragraph, the less, the reviews. Or the less real, it is?\\

Person 1: No, no, no. Uh, you, you the opposite. Yeah. Okay. It, it would be realistic that if it has more objectives, something like that.\\

Person 7: Okay. Okay. Well, uh.\\

Person 1: Oh, what do you mean by adjective? Like, do you mean the choice of adjective or?\\

Person 7: No, just any adjective. So for example, the ones that I have rated the, the, the realest, have more objectives than, than the other ones. So it could be just, um, your personal opinion, it's not about.\\

Person 1: Okay.\\

Person 7: Could you tell me why you think having more adjective make it more realistic?\\

Person 1: Just coincidence, I mean, I just read it and then I just, uh, uh, ranked it.\\

Person 7: But generally, in fake review, generally people say more, like a lots of terms of objectives, "This is good. This is best. This is something I never imagined. This is, uh, like cost-effective, blah, blah, blah, blah, blah, blah," so, generally, this is the case. Anyway, do you, um, like, uh, the questions, uh, to both of you, do you think the explanations, um, that we call explanations, the highlighting the word that is relevant to the decisions, do you think these explanations of the highlighted words, uh, helps you to refine your thoughts that whether is fake or not, and like how helpful that viewers, uh, explanation is?\\

Person 6: For me, for me, they didn't really help me to find out if they are or not real. Uh, I think it's more like a context thing. Only, I mean for me the word has, has to, um, it's okay. It was the, the same.\\

Person 1: How about you? Okay. Okay.\\

Person 7: Yeah, I just pay attention to whole the review.\\

Person 1: Okay, not a specific words. It's about the context and the, like linguistic structures and all this thing? Okay, okay. Yeah, yeah, yeah. Okay. Yeah.\\

Person 7: Yeah. Yeah. Exactly. Romantic. Dramatic. Dramatic, dramatic, dramatic. Yeah, that's it. Okay. Yeah. Dramatic, dramatic. Not romantic.\\

Person 1: Um, so I gave the first review.\\

Person 7: Um, it would be good if you say a bit louder. Okay.\\

Person 1: I gave the first review one star because, um, you know, it's fake because I think it is really unlikely to happen that a child read a book and read it all over again. So it's not about the language that he use, it's more about, um, that it's a little bit overtaken, the review. Um, the second one I also highlight the sentence that are repeated. And I think repetitions, um, are more likely for AI recommendation. So that's why I give this one also one star, and the third one actually was from me, at the first, um, I gave them five stars. So that it's likely to happen because it's short. I think in real life everyone just give short recommendations and not long recommendations. But then after reading the AI, um, explanation, yes. Um, reading the words excellent, amazing, great. I also think that it's, it's not a real review.\\

Person 1: Okay, fake, okay? Okay.\\

Person 7: Okay.\\

Person 1: Okay, so, so you found, uh, the explanations for, uh, review three is helpful to refine your decisions because, uh, you spotted some words that was not like, uh, you didn't think in the plain text. Okay.\\

Person 7: Yes, yes. Definitely. In the first place. Yes. Yes.\\

Person 1: Um, the words were excellent, amazing, highly recommended. So that's more likely for AI. And at the first, when I read this, I didn't.\\

Person 7: Make sense of those words there, there. Okay. Yes, yes. Yes. Yes. Yes.\\

Person 1: So you, you think, uh, with the explanations you find it that these are exaggerating, uh, uh, of some words and recommendations.\\

Person 7: It doesn't, so, so it doesn't feel like something you would type. Is that what you mean? That those words that, those highlights makes you like kind of read it again? Is that what you mean? Or more like.\\

Person 7: Uh, those highlights were more, I think it's more unlikely that a person, um, write a review that, um, describes a product excellent. So it's more overrated. Unlikely.\\

Person 1: Okay. Okay. Okay. Good.\\

Person 7: Okay.\\

Person 1: Thank you. Yeah. Yeah. Thank you so much. That's good. Yeah, that's good. Yeah. Yeah, what's your thoughts?\\

Person 8: Um, firstly, I think this, for me, at the first, first place, I think this one, the third one is a, is a real review. Um, just because it's.\\

Person 1: Is a real or maybe real or, how do you.\\

Person 8: I think maybe real. Yeah, maybe real.\\

Person 1: Okay, maybe real, okay?\\

Person 8: Um, just because it's.\\

Person 1: It could be a bit louder. Okay.\\

Person 8: Okay. Just because it, it's a bit short and it also has the name on it. Probably they know someone from the, um, cafe or whatever. So I think that is pretty real. And the, and also that AI didn't affect or much as well. I think what I find.\\

Person 1: Did the, uh, AI explanations at least help you to, to make understand, okay, um, uh, these are the words that is relevant. That this is, this could be fake or real or something like that?\\

Person 8: Yeah, I think the, that, even the words make me feel like probably it's real. It's not fake. Even the word finding on this.\\

Person 1: So you find the explanations, uh, meaningful or meaningless or, like how do you rate it out of 10?\\

Person 8: Meaningless for the third one.\\

Person 1: Okay.\\

Person 8: Yeah. But the second one, um, this one I find it's, it's probably, it's definitely fake. People wouldn't give such personal information on this. And also such a long message regarding review on the gloves. And that's weird. So I think this one is, um, is totally fake and also these words also helped me to find out, or to, to stronger my opinion. I think, maybe, yeah, this is fake.\\

Person 1: Okay. So, so it helps you, uh, for this review two? Okay. Okay. And then how, how do you rate for this one?\\

Person 8: Yeah. Yeah. Definitely. Yeah.\\

Person 8: Yeah, this one it's, um, it's, it doesn't make any sense for these, um, sentences. Yeah. And also the finding is also pretty much the same as the word I find. That, um, I find, I think this one is fake as well. Yeah.\\

Person 1: Okay. Okay. Yeah. Yeah, thank you so much. So this participants said reviewer three, and then two, and one.\\

Person 8: Thank you.\\

Person 1: Okay. Yeah, yeah, yeah. Yeah, yeah, yeah, yeah. Thank you. Yeah. Okay. Bob, the first one you gave, uh, two out of.\\

Person 9: Can you say a bit louder, I think. Okay. Uh, this is the first text for review. So I review it and, um, I gave, uh, two out of five. And before I rate, uh, I have positive, positivity back about it.\\

Person 1: So this is two out of five or three out of five?\\

Person 9: Yeah.\\

Person 1: The first one, I saw that it's three out of five and then I. Okay, then two out of five.\\

Person 9: The first one, I told that it's three out of five then I changed it to, yeah. First of all, I had positive idea about it because, uh, he said that, I had it read before, still my favorite book, um, this interaction with kids, he liked. Uh, can't put it down, a chance, uh, just if he had chance to take it to college, it gave me positive, um, vibe actually. But after that, uh, at last, he said that, uh, he wanted to know more about the characters. Just, uh, it seem to me that it's fake, because before he said that, uh, "I read it over and over again." But at last, he said, uh, "I want to know more about characters." So that's, uh, seemed to me fake. So that's why I just do that. And, uh, at last about the AI, uh, actually nothing changed for me, uh, with AI help, uh, just, uh, from the first insight I found the mistake there. So nothing changed with AI explanation, and.\\

Person 1: So you haven't find that this is something that make your sense. Okay. Okay.\\

Person 9: No, no, no. And, um, this is the last one or second one?\\

Person 1: Yeah.\\

Person 9: So about the second one, uh, just, um, first of all, I saw that it might be half, about zero out of five. I didn't believe it because he repeated and exaggerated all the words. He did all the words again and again. And at last, he says that he didn't use them. He says, just praised it. And at first, he says that I actually didn't use them. It's also fake and nothing changed with the AI explanation, because I found it.\\

Person 1: Five, okay? Yeah.\\

Person 9: Yeah. Use them. Yeah. So.\\

Person 1: Yeah. So yeah, you, you, you found it, but do you find this helpful? No? Okay.\\

Person 9: About the first one. No, it's, no, it's not helpful for me. And about the third one, uh, I believe it actually, it's more emotional. It's just easy explanation and no repetition. Uh, it gave me just a not just real vibe, because it's, uh, emotional, it should be against with, "Oh my God, this place highly." Something like that. And actually, for the third one also, uh, for the third one also nothing has changed for me with AI explanation. Just fine itself.\\

Person 1: Okay.\\

Person 9: Yeah. Thank you so much. Yeah. It's my pleasure and it's your time. Uh, it's the last one.\\

Person 1: Thank you. Thank you. So. What about you, yeah? So the first one, review one. You can start with review one. This is one. This is two, this is three.\\

Person 10: Okay. So, uh, the first one, I guess, um.\\

Person 1: You can say a bit louder.\\

Person 10: Yeah, some words were a little bit like AI generated, like avid reader. I don't think that people in general really use that avid word. But in general, it kind of looks real. But for.\\

Person 1: So first one you, you, you think it's a kind of real?\\

Person 10: Um, not that much of a real, but maybe, um, they have even helped, uh, from, I don't know maybe, uh, some ChatGPT to rewrite it as well. Maybe that, uh, that could be done, uh, but, uh, the AI, um, explanation for me wasn't really that much helpful because, uh, I don't know, is it like about just the words, or is it about the context and structure as well? Because I see only words here. And you can't.\\

Person 1: Okay.\\

Person 1: Okay.\\

Person 1: Yeah. Here in this model the, the model, uh, only think that this words could have an influence to be fake or to be, uh, authentic.\\

Person 10: Yeah, but tell us about the structure you want to say.\\

Person 1: Because, um, the structure is like the most important thing when you wanna see that if anything is AI generated as well. Because like the words like, I can use these words if I want. Uh, like you can, you can't just decide on words. I believe.\\

Person 1: Okay. Okay. Yeah, yeah, yeah. Yeah.\\

Person 10: Yeah.\\

Person 1: Could you tell us like how, you, how do you think the AI would structure such a fake review?\\

Person 10: Well, for example, the second one, I, I think that, um, either there's like an illiterate man who's been paid to talk about this, or he's just like a really bad AI that they just like, uh, you can just, like, uh, write some sentences and then give it to AI and tell them to rephrase. If it's like, uh, not a really good AI, it will rephrase it something like, um, with really bad words, with, uh, some repetitive words, exaggerating, is something like that. But also for the third one, um, the thing that was much, much more human-like for me was, uh, the grammatical errors and some errors that, for example, when she was about to type, maybe she missed a letter. I mean, I don't think that AI does that.\\

Person 1: Okay.\\

Person 10: Yeah. It's my opinion. Based on my experience.\\

Person 1: Based on your experience with ChatGPT?\\

Person 10: Yeah. Okay, then, uh, so you think AI would have like a more, like it would give more context and have like a more unnecessary structure.\\

Person 1: So you think, uh, so you think AI would have like a more, like it would give more context and have like a more unnecessary structure.\\

Person 10: Yeah, but not, not more human-like structure, you know? Like, it's not something that I can say it, but it's something that you feel it.\\

Person 1: And then human will be more, uh. Yeah, yeah.\\

Person 10: Yeah.\\

Person 1: Feel like structure. Okay. Okay. Thank you.\\

Person 10: Yeah, okay, okay. Thank you so much.\\

Person 1: Would you say AI is more logical than human? Like, what, what, what the difference? What do you mean by human structure?\\

Person 10: Like, uh, no, it's not more logical, but it's, it's complicated. I don't know how to say it.\\

Person 1: Don't, don't worry. Don't worry. That's also hard.\\

Person 10: No, but I don't mean that it's more logical. I, I mean that it's more robotic form. Like, you know, um, sometimes when you read something you realize that, okay, yeah, this is from a robot because it's, uh.\\

Person 1: Okay. Yeah, yeah.\\

Person 1: Because the, the logic doesn't link to each other, maybe?\\

Person 10: Yeah, maybe. Yeah. And, yeah. That's, that's it also.\\

Person 1: Yeah. Okay. Yeah, yeah. Thanks. Thank you. Thank you. Now, uh, we are going to.\\

Person 10: Yeah.\\

Person 1: Yeah, yeah.\\

Person 11: Well, I want to ask one last question. Is, is this because, uh, the structure is more interpretable for you then it becomes more human-like? And then for AI is like, okay, why are you suddenly talk about this, and why are you suddenly talking about this? Why do you start with this?\\

Person 10: But also, some AIs are really that much strong that they don't really go to this and that.\\

Person 11: Yeah. Yeah, it's scattered.\\

Person 10: Uh, I, I use some AI for some texts and I see that some of them are really weak. Some of them are really, really bad. But some of them are actually really strong that you may think that, okay, maybe some professor, uh, wrote that. Yeah, because so, that's why that I think that AI explanation, the ML explanation here by the words itself, it's definitely not enough. And it's maybe, maybe even not enough for me, because right now AI is getting much, much stronger every day. But in general, yeah, that was I guess.\\

Person 1: Yeah. Yeah. Yeah, yeah.\\

Person 11: Okay, okay.\\

Person 1: Yeah, yeah. Okay, good. Could you pause the recording?\\

Person 10: Okay.\\

Person 1: Seema, what do you think?\\

Person 12: Okay.\\

Person 1: Yeah. She is also our participant.\\

Person 12: Yeah.\\

Person 1: Uh, start with the, uh, review one.\\

Person 12: For review one, I feel it is like, uh, three stars and feel it, it could be human because, uh, because it is talking about books and, uh, books you can, uh, you can have exaggeration of the feelings, because, uh, maybe it is the feelings of nostalgia. But at the same time, I feel it's the, the high level of exaggeration. And the books is talking about from childhood to the, uh, after college. It's not the book, people are not reading the same book during this, uh, process. Because after I feel maybe it is not human, but when I, uh, read them keywords. The keywords are the normal keywords, like chance, character. No. And I feel maybe, uh, because of the normal keywords that provide me, um, maybe it could be more human, but still, I feel it is, it is fake. For the review number two, I feel, the way it's talking about gloves, nobody talk about gloves like this today. And, um, it's like, um, there is no nostalgia, there is no feeling, there is no, and nobody, uh, want to have, be happy with gloves, you know what I mean? It's talking about, "Oh, you can be really happy with this glove." Nobody care about the happy feeling of gloves.\\

Person 1: Okay.\\

Person 1: It could be. Okay.\\

Person 1: Yeah.\\

Person 1: Okay.\\

Person 1: Yeah. Yeah. Yeah. Yeah. Yeah.\\

Person 1: Okay. Overreacting. Yeah\\

Person 12: The way it's talking is, yeah, it's really, really, really exaggeration to talk about glove, a glove. And it is not the way people express their feeling. They can say, "Okay, it was, it works, it's okay." And nobody talk like this long, uh, conversation just to express their feeling about gloves. For the third one, I feel it is, uh, like, uh.\\

Person 1: Yeah. Yeah.\\

Person 12: Human.\\

Person 1: Human.\\

Person 12: Yes, and, um, it was about travel, yeah. I think all of us may experience the travel that I feel it was fabulous, it was perfect, it was great. And all of us maybe have a chance to go to travel for our birthday, and then feel, okay, it was a great birthday because of the travel, because of the location that I visited, because of the service that I was received. This is real. And, um, also it is, I don't know how to say.\\

Person 1: So I can see that, uh, uh, you changed, uh, to a real review.\\

Person 12: One more star. Yeah.\\

Person 1: So yeah, so, uh, is the, uh, explanations helps you to?\\

Person 12: Yeah. Yeah. It is, because I feel it focus on birthday. And it is, it can be like excellent birthday, amazing, and then food. The, the words it is expressing I feel they are real words and they are the words that we are using as a human. Maybe the way you are saying, "Oh, that was, that, that food was perfect. That travel was, uh, unforgettable." This is real, yeah.\\

Person 1: Yeah. Yeah.\\

Person 1: Yeah. Yeah. Yeah. Thank you so much.\\

Person 12: You're welcome.\\
\fi
\end{document}